\documentclass[10pt,twocolumn,letterpaper]{article}
\usepackage{titlesec}
\usepackage{algpseudocode}
\usepackage{multirow}
\usepackage{amsmath,amssymb,amsfonts}
\usepackage{graphicx}
\usepackage{textcomp}
\usepackage[section]{placeins}
\usepackage{algorithm}
 \usepackage{algpseudocode}
 \usepackage{algorithmicx}
 \usepackage[accsupp]{axessibility}
 \algdef{SE}[DOWHILE]{Do}{doWhile}{\algorithmicdo}[1]{\algorithmicwhile\ #1}%

\algrenewcommand\algorithmicrequire{\textbf{Require:}}
\algrenewcommand\algorithmicensure{\textbf{Output:}}
\usepackage[pagenumbers]{wacv} 
\usepackage{graphicx}
\usepackage{amsmath}
\usepackage{amssymb}
\usepackage{booktabs}

\def \bbR{{\mathbb{R}}}

\def \cC{{\mathcal{C}}}

\def \cE{{\mathcal{E}}}
\def \cI{{\mathcal{I}}}

\def \cN{{\mathcal{N}}}
\def \cT{{\mathcal{T}}}

\usepackage[pagebackref,breaklinks,colorlinks]{hyperref}

\usepackage[capitalize]{cleveref}
\crefname{section}{Sec.}{Secs.}
\Crefname{section}{Section}{Sections}
\Crefname{table}{Table}{Tables}
\crefname{table}{Tab.}{Tabs.}

\begin{document}

\title{ Dataset Augmentation by Mixing Visual Concepts}

\author{
Abdullah Al Rahat, Hemanth Venkateswara\\
Georgia State University,  USA\\
{\tt\small mkutubi1@student.gsu.edu, 
hvenkateswara@gsu.edu}
}
\maketitle

\begin{abstract}

This paper proposes a dataset augmentation method by fine-tuning pre-trained diffusion models. 
Generating images using a pre-trained diffusion model with textual conditioning often results in domain discrepancy between real data and generated images. 
We propose a fine-tuning approach where we adapt the diffusion model by conditioning it with real images and novel text embeddings. 
We introduce a unique procedure called Mixing Visual Concepts (MVC) where we create novel text embeddings from image captions. 
The MVC enables us to generate multiple images which are diverse and yet similar to the real data enabling us to perform effective dataset augmentation. 
We perform comprehensive qualitative and quantitative evaluations with the proposed dataset augmentation approach showcasing both coarse-grained and fine-grained changes in generated images. 
Our approach outperforms state-of-the-art augmentation techniques on benchmark classification tasks. The code is available at \href{ https://github.com/rahatkutubi/MVC}{ https://github.com/rahatkutubi/MVC}
\end{abstract}

\section{Introduction}
\label{sec:intro}
Deep Neural Networks (DNNs) have achieved significant success due to their ability to learn complex representations from large datasets. However, they often require extensive data to prevent overfitting, which poses challenges in domains like visual recognition or medical imaging, where creating and annotating data such as MRI and X-ray images can be costly and time-consuming. Data augmentation is a key solution to this problem.

 Data augmentation is used to increase the dataset size by generating new samples. Traditional image augmentation methods include geometric transformations (e.g., translation, rotation, affine transformations), color space transformations, kernel filters (e.g., blur, smoothing, sharpening), and image cropping \cite{chlap2021review}. 
Data augmentation techniques like AutoAugment \cite{cubuk2019autoaugment}, RandAugment \cite{cubuk2020randaugment},
and geometric transformations of data are now an integral part of data pre-processing for training deep neural networks, and help in avoiding overfitting. 
While these methods effectively increase dataset size, they lack natural variation as they are based on modifying existing data without introducing new concepts.

\begin{figure}[tb]
  \centering
\includegraphics[height=6.5cm,width=8.4cm]{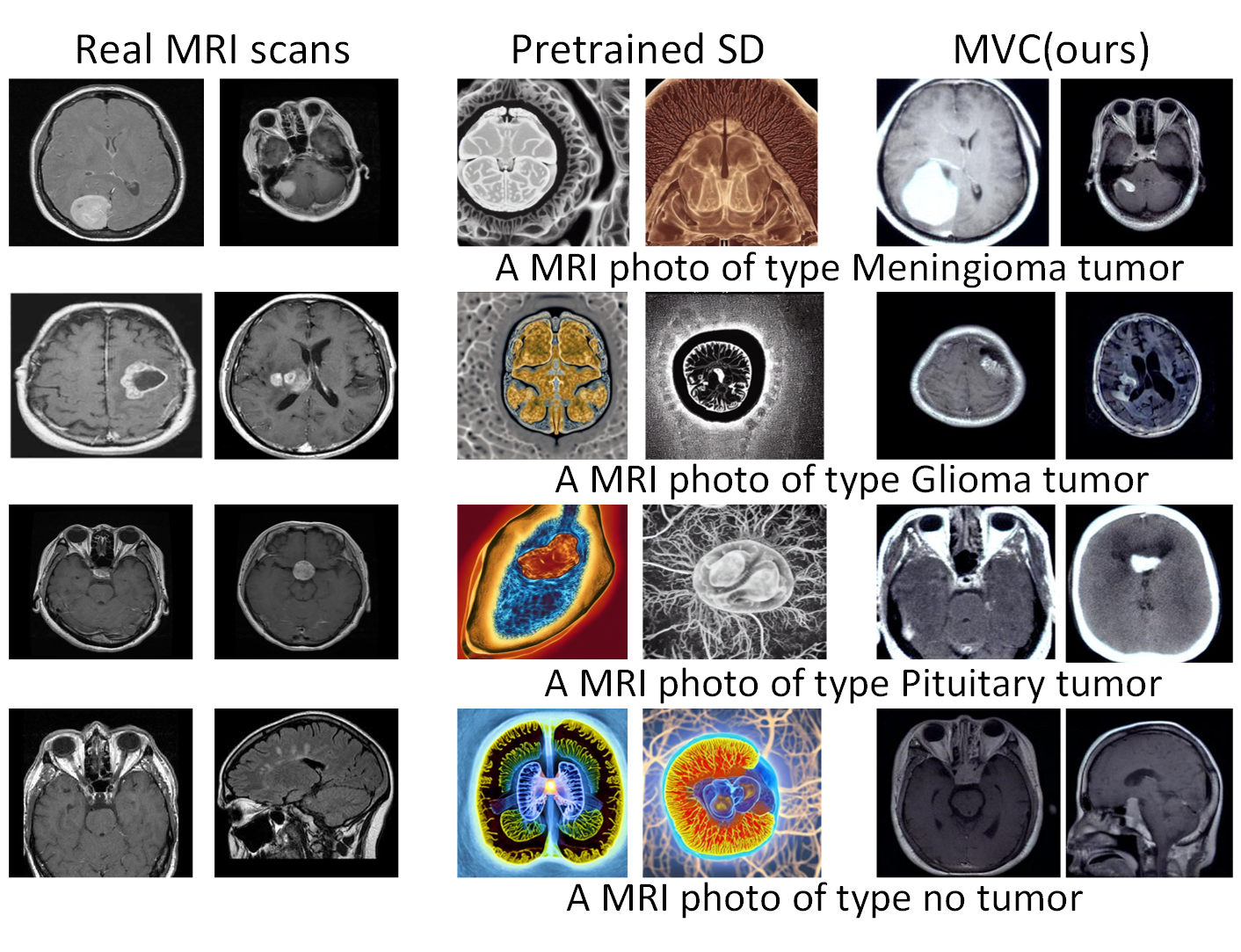}
\vspace{-25pt}
\caption{Na\"ively deploying a pre-trained generative model to generate new images for dataset augmentation can lead to domain discrepancy. Columns 1, 2 are MRI scans from the Brain Tumor Dataset \cite{1jny-g144-23}, Columns 3, 4 are images generated using a pre-trained Stable Diffusion (SD) model, Columns 5, 6 are images generated using the proposed MVC method.}
\label{Fig:Fig1}
\vspace{-20pt}
\end{figure}

 Recent advancements in deep generative models, such as Variational Autoencoders (VAE) \cite{kingma2013auto}, Generative Adversarial Networks (GAN) \cite{goodfellow2014generative}, Normalizing Flow \cite{rezende2015variational}, Autoregressive Models \cite{van2016conditional}, and Diffusion models \cite{ho2020denoising} have opened new avenues for generating synthetic data. These models can produce high-quality outputs such as realistic images, natural language, and diverse music and speech \cite{dhariwal2020jukebox}.
Among these, diffusion models have emerged as the most versatile approach to generate highly detailed, photorealistic, and diverse samples, surpassing GANs and other models in quality and diversity. The rapid growth in research on diffusion models over the past two years highlights their potential for high-quality, diverse sample generation \cite{croitoru2023diffusion}. Notable examples include DALL-E 2 \cite{ramesh2022hierarchical}, Imagen \cite{saharia2022photorealistic}, and Stable Diffusion \cite{rombach2022high}, which have gained significant attention for their high-quality text-to-image generation capabilities.

 Recent studies have demonstrated the effectiveness of synthetic images generated by generative diffusion models in classification tasks \cite{he2022synthetic}. However, relying solely on pre-trained diffusion models that are conditioned by textual prompts from large language models can result in domain discrepancies between the real and generated data. 
In order to augment a dataset we propose to use image captions that can condition the diffusion model to generate images similar to the images in the dataset. 
This process is constrained to generate as many new images as the number of captions. 
But, can we not use a language model to generate additional similar captions? Yes, but that leads to domain discrepancy between generated and training data due to the variations in the new captions introduced by the language model. 
We therefore propose to work with caption embeddings rather than captions. 
In this work we introduce the technique of Mixing Visual Concepts (MVC) where we generate additional caption embeddings by mixing CLIP \cite{radford2021learning} generated embeddings of existing captions. 
We gain in three ways; (1) we are able to generate a large number of images, (2) the generated images are similar to the training data, and (3) we can control the diversity of the images by controlling the extent of mixing. 
Figure \ref{Fig:Fig1} compares the generated images produced by a pre-trained diffusion model and a fine-tuned diffusion model. It is evident that applying a pre-trained model na\"ively without adaptation can lead to domain discrepancy in the augmented dataset.

Inspired by InstructPix2Pix \cite{brooks2023instructpix2pix}, we condition a Stable Diffusion model \cite{rombach2022high} with novel caption embeddings to generate new images. 
We begin by generating captions for the images in our dataset using a vision-language model \cite{li2023blip}. 
We apply the pre-trained CLIP to extract embeddings for these captions. 
We propose the MVC algorithm to generate novel embeddings by \emph{mixing} the CLIP-generated embeddings. 
We fine-tune a pre-trained Stable Diffusion model \cite{rombach2022high} conditioned with novel caption embeddings along with the images in the dataset to generate new images for dataset augmentation. 
Our procedure generates images with distributional consistency and natural variations similar to real images. 
Our main contributions are as follows: (1) We introduce a robust dataset augmentation technique to generate within-domain data, (2) We demonstrate an efficient procedure to generate novel embeddings by mixing visual concepts (MVC) to condition the generative model, (3) We conduct extensive experiments to validate the effectiveness of the generated images on downstream classification tasks.



\section{Related Work}
Data augmentation techniques for image recognition are typically created manually, and the most effective methods are often tailored to specific datasets. For instance, elastic distortions, scaling, translation, and rotation are commonly used for the MNIST dataset by top-performing models \cite{rezende2014stochastic,kingma2019introduction,ngiam2011learning}. Meanwhile, random cropping, mirroring, and color shifting or whitening are more frequently employed for natural image datasets like CIFAR-10 and ImageNet \cite{russakovsky2015imagenet}. However, these techniques require significant time and expert knowledge to develop \cite{cubuk2018autoaugment}.\\

 There has been a notable shift in the field of data operations, with researchers now focusing on enhancing data by combining various strategies. For example, Smart Augmentation merges multiple samples from the same class to create innovative data networks \cite{lemley2017smart}. Meanwhile, Tran et al. \cite{tran2017bayesian} utilize a Bayesian approach to generate augmented data based on the distribution learned from the training set. DeVries et al. \cite{devries2017dataset} use noise, interpolations, and extrapolations in the feature space to enhance data.\\

 In recent years AutoAugment \cite{cubuk2019autoaugment} and RandAugment \cite{cubuk2020randaugment}, have been extensively applied for dataset augmentation. 
They have been instrumental in achieving impressive results in image classification, self-supervised learning \cite{kim2016deep,gui2021review,rezende2015variational}, and semi-supervised learning \cite{bond2021deep} on the ImageNet \cite{russakovsky2015imagenet} benchmark.  These augmentations are highly versatile and are particularly effective at encoding invariances to data transformations, making them a valuable tool for image classification. Methodologically, these find the best policies to group the primitive transformations to apply to real images in order to generate augmented images.\\

For image recognition tasks, additional datasets\cite{dosovitskiy2015flownet,peng2017visda,richter2016playing} are usually synthesized from a traditional simulation pipeline that uses a specific data source, such as synthetic 2D renderings of 3D models or scenes from graphics engines \cite{he2022synthetic}.
However, this approach has some drawbacks, including the potential for a discrepancy between distributions of the synthetic and real-world data, the need for large physical storage space, the high costs for sharing and transferring, and limitations in data amount and diversity due to using a specific data source \cite{he2022synthetic}. On the other hand, generative models represent a more efficient approach to generating synthetic data, capable of producing high-fidelity, photorealistic images in large numbers. \\

Examples of attempts to use synthetic data from generative models for image recognition include a class-conditional Generative Adversarial Network(GAN) \cite{besnier2020dataset} to train classifiers for the same classes, and producing labels for object part segmentation \cite{zhang2021datasetgan} from the latent code of StyleGAN \cite{karras2019style}. 
Although these works have shown promising results, they are limited in scope and focus on specific tasks \cite{he2022synthetic}. On the other hand, Jahanian et al. \cite{jahanian2021generative} used a GAN-based generator to produce multiple views for unsupervised contrastive representation learning. Shifting away from GAN-based augmentation, recent studies \cite{shipard2023boosting,trabucco2023effective} have highlighted the ability of diffusion models to generate training data with minimal or no examples, producing realistic training samples \cite{jain2022distilling}. However, models trained solely on diffusion-generated data typically perform worse compared to those trained on real datasets, unless specifically fine-tuned for the target task \cite{azizi2023synthetic,trabucco2023effective}. 
This is due to the domain gap between synthetically generated data and real data. \\

Our study aims to create augmentations that maintain visual consistency with the original training data. To achieve this, we explore blending micro concepts or characteristics from multiple images into generated images using engineered prompts in the CLIP \cite{radford2021learning}  embedding space. 

\section{Method}

\subsection{Background and Preliminaries}

\textbf{Diffusion Model}: A Denoising Diffusion Probabilistic Model (DDPM) \cite{ho2020denoising} is a generative model that is designed to reverse (denoise) a diffusion process. 
The diffusion process gradually adds varying amounts of Gaussian noise $\epsilon \sim \cN(0,I)$  to an input image $x_0$ over a specified number of timesteps $t\in\{1,2,\ldots,T\}$ until the resulting image is totally Gaussian noise. 
A diffused image $x_t$ is obtained from input image $x_0$ based on the following equation, 
\begin{equation}
x_t = \sqrt{\bar{\alpha}_t} x_0 + \sqrt{1 - \bar{\alpha}_t}\epsilon,
\label{Eq:diff1}
\end{equation}
 where $\bar{\alpha}_t = \prod_{i=1}^t\alpha_i$, and $\alpha_t$ is a time dependent noise scaling factor with $\alpha_1 = 1$ (no noise), and $\alpha_T \approx  0$ (only noise). 
The diffusion model is a denoiser neural network with parameters $\theta$.
It is trained to predict the noise $\epsilon$ diffused in the image. The parameters $\theta$ are estimated minimizing the following objective: 
\begin{equation}
L =  \min_{\theta} \mathbb{E}_{x_0, \epsilon,t} \| \epsilon - \epsilon_{\theta}(x_t, t) \|_{2}^{2}.
\label{Eq:diff2}
\end{equation}
Starting with pure noise $x_T\sim \cN(0,I)$, the denoiser network $\epsilon_\theta(x_t,t)$  predicts the noise in $x_t$. The noise is subtracted from $x_t$ to transform it to $x_{t-1}$. This iterative procedure for $T\geq t \geq 1$ eventually yields the generated image $x_0$.\\

\textbf{Latent Diffusion Model (LDM)}: The LDM is an efficient diffusion model that performs diffusion and denoising in a reduced-dimensional latent space rather than in the high-dimensional pixel space \cite{rombach2022high}. 
A LDM  trains an Autoencoder $D(E(x))$ to map an image $x$ to a latent space, where $z \gets E(x)$ is the latent representation and $E(.)$ is the Encoder. 
The Decoder $D(.)$ recreates the original image $\hat{x}\gets D(z)$. 
In addition, the denoising process can be conditioned to generate a desired image. 
The LDM is implemented as a time-conditioned U-Net \cite{ronneberger2015u} where cross-attention \cite{rombach2022high} is applied to integrate conditioning information into the denoising process to generate desired images. 
Conditioning information can be input in different modalities. For e.g. class labels, text prompts, segmentation maps, etc., are embedded into fixed-dimensional representations and input to the LDM denoiser to guide the denoising process. 
The objective function to train the LDM is,  
\begin{flalign}
L_{LDM} &=   \min_{\theta} \mathbb{E}_{z_0, \epsilon, t}\| \epsilon - \epsilon_{\theta}(z_t, t, e_{\cT}) \|_{2}^{2},
\label{Eq:diff3}
\end{flalign}
where $e_{\cT}$ is the embedded conditioning. 

In case of dataset augmentation, we aim to keep the generated images similar to the images in the dataset. Inspired by the work in InstructPix2Pix \cite{brooks2023instructpix2pix}, we modify the U-Net architecture in the LDM to leverage additional information in the form of image embedding $e_{\cI} \gets E(x)$. 
These embeddings are concatenated to the noisy latents $z_t$ to condition the denoising process to generate images similar to $x$. 
Accordingly, we update the objective function as follows:
\begin{flalign}
L_\text{LDM} &=  \min_{\theta} \mathbb{E}_{z_0, \epsilon, t} \| \epsilon - \epsilon_{\theta}(z_t, t, e_{\cT},e_{\cI}) \|_{2}^{2}. 
\label{Eq:LDM}
\end{flalign}
During image generation, we first embed the noise image $x_T \sim \cN(0,I)$ into the latent representation $z_T \gets E(x_T)$. 
The denoiser network $\epsilon_{\theta}(z_t, t, e_{\cT},e_{\cI})$ iteratively predicts the noise to transform $z_t$ to $z_{t-1}$ in the following manner, 
\begin{flalign}
z_{t-1} &= \frac{1}{\sqrt{\alpha_t}}\Big(z_t - \frac{1-\alpha_t}{\sqrt{1-\bar{\alpha}}_t}\epsilon_{\theta}(z_t, t, e_{\cT},e_{\cI}) \Big).
\label{Eq:denoise}
\end{flalign} 
Arriving at $z_0$, the decoder is applied to yield the generated image $\tilde{x} \gets D(z_0)$. 
The text prompt embedding $e_{\cT}$ and the image embedding $e_{\cI}$ condition the generation process to yield desired images. \\

\textbf{Classifier-free guidance}: One of the key challenges in text-guided generation is the amplification of the effect induced by the conditioned text. To this end, Ho et al. \cite{ho2022classifier} have presented the classifier-free guidance technique, where the denoising is also performed unconditionally, which is then extrapolated with the conditioned prediction. More formally, let $\emptyset = \psi(\text{`'})$ be the embedding of a null text and let $w$ be the guidance scale parameter, then the classifier-free guidance prediction is defined by:
\begin{flalign}
{\epsilon}_{\theta}(z_t, t,e_{\cT},e_{\cI}, \emptyset) =& w \cdot \epsilon_{\theta}(z_t, t, e_{\cT},e_{\cI}) \notag \\
&+ (1 - w) \cdot \epsilon_{\theta}(z_t, t, \emptyset),
\end{flalign}

\noindent where $w = 7.5$ is the default parameter for Stable Diffusion. We will now outline the procedure to create multiple embeddings by randomly mixing visual concepts from a limited set of embedded representations. 

\subsection{Mixing Visual Concepts}
Consider all the $K_y$ images $\{x^y_k\}_{k=1}^{K_y}$ belonging to a category $y$ in the training dataset. 
It is our goal to generate $N_y$ new images $\{\tilde{x}^y_k\}_{k=1}^{N_y}$ all with the same label $y$. 
We wish the generated images to be unique and yet `similar' in style to the real images. 
We propose to achieve this by mixing visual concepts from multiple real images to generate a new image. 
We leverage the CLIP model which generates high-quality latent representations that effectively bridge visual and textual data \cite{radford2021learning}.\\ 
\vspace{-10pt}
\begin{algorithm}[H] 
    \caption{ Mixing Visual Concepts}\label{Alg-Decap}
    \begin{algorithmic}[1]
        \Require{ $\cC:=\{c_k\}_{k=1}^{K_y}$: Set of $K_y$ Captions for category $y$, and $\texttt{CLIP}(.)$: CLIP Text Embedder
}
            \State $\cE :=\{e_k\}_{i=k}^{K_y}$ where $e_k \in \bbR^{m\times d}$ and $e_k\leftarrow \texttt{CLIP}(c_k)$
            \State $\hat{\cE} := \varnothing$ \quad \texttt{\scriptsize{//set of mixed CLIP embeddings}}
        \For{$k \gets 1$ to $N_y$}
            \State $\tilde{e}_k \leftarrow \texttt{RandSample}\{\cE\}$
            \For {$p \gets 1$ to $P$} \quad \texttt{\scriptsize{//Coarse mixing}}
                \State $e_p \leftarrow \texttt{RandSample} \{\cE \setminus \{\tilde{e}_k\}\}$
                \State $r,s \leftarrow \texttt{Uniform}\{1,2,\ldots,m\}~\text{with}~ r<s$
                \State $[\tilde{e}_k]_{r:s,:} \leftarrow [e_p]_{r:s,:}$ ~~ \texttt{\scriptsize{//replace rows $r:s$}}
            \EndFor
            \For {$q \gets 1$ to $Q$} \quad \texttt{\scriptsize{//Fine mixing}}
                \State $e_q \leftarrow \texttt{RandSample} \{\cE \setminus \{\tilde{e}_k\}\}$
                \State $u,v \leftarrow \texttt{Uniform}\{1,2,\ldots,d\}~ \text{with}~ u<v$
                \State $w \leftarrow \texttt{Uniform}\{1,2,\ldots,m\}$
                \State $[\tilde{e}_k]_{w, u:v} ~~ \leftarrow [e_q]_{w, u:v}$  \texttt{\scriptsize{//replace $u:v$ in row $w$}} 
            \EndFor
            \State $\tilde{e}_k \gets \tilde{e}_k ~\text{\textcircled{c}}~ \texttt{CLIP}(\varnothing)$ \quad \texttt{\scriptsize{//Concat Class-free guidance}}
            \State $\hat{\cE} \gets \hat{\cE}\cup\{\tilde{e}_k\}$
        \EndFor
    \end{algorithmic}
    \label{Alg:1}
\end{algorithm}

 \begin{figure}[tb]
  \centering
  \includegraphics[height=6cm, width=8.4cm]{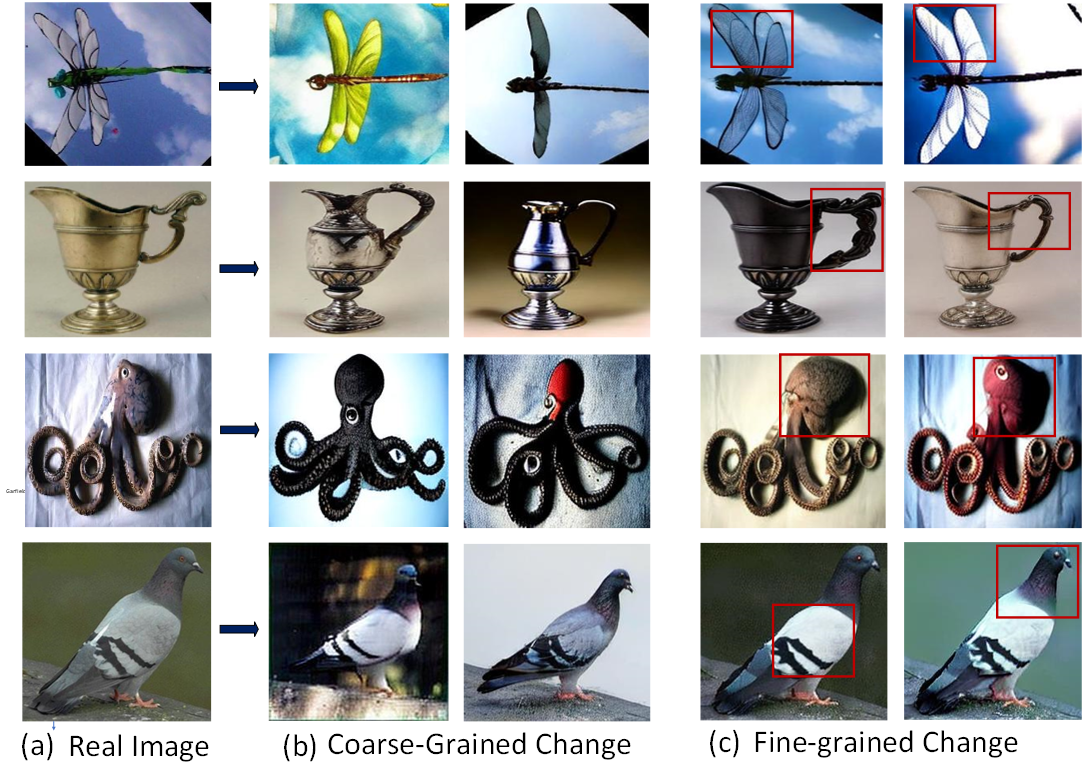}
  \vspace{-15pt}
  \caption{The results of coarse and fine mixing. (a) real image,  (b) generated images conditioned on the real image along with coarse mixing, (c) generated images conditioned on the real image along with only fine mixing.
  }
  \label{Fig:coarse}
  \vspace{-15pt}
\end{figure}

We begin with getting textual description/captions for the images using a pre-trained BLIP-2 model \cite{li2023blip}. 
We prefix each caption with the sentence, ``\texttt{This is an image of <y>}'' to ensure the category is captured in case the BLIP-2 model misses to label the image in the caption. 
This generates a set of captions, $\{c^y_k\}_{k=1}^{K_y}$. 
We arrive at the latent representations $\{e^y_k\}_{k=1}^{K_y}$ where $e \gets \texttt{CLIP}(c)$ and $e \in \bbR^{m \times d}$. 
The CLIP embedding has a context of $m$ tokens where each token is embedded in $d$ dimensions. 
We propose to create new embeddings by mixing tokens from pre-existing embeddings. 
Starting with an embedding $e_i$, we replace some of its rows with corresponding rows from another embedding $e_j$. 
We term this \emph{coarse mixing} when we make a relatively big edit to an embedding. 
We attempt to make fine changes to an embedding by replacing a few elements in the embedding with corresponding elements from other embeddings. 
We term this \emph{fine mixing}. 
Algorithm \ref{Alg:1} outlines the procedure to generate a set of new CLIP-like embeddings $\hat{\cE}$ from pre-existing embeddings. 
Figure \ref{Fig:coarse} displays the generated images with Coarse and Fine mixing.

\subsection{Training and Generation}
The new embeddings in $\hat{\cE}$ are used to condition a diffusion model to generate new images similar to the images in the dataset. \\

\begin{figure}[!t]
  \centering
  \includegraphics[height=7.8cm, width=9.7cm]{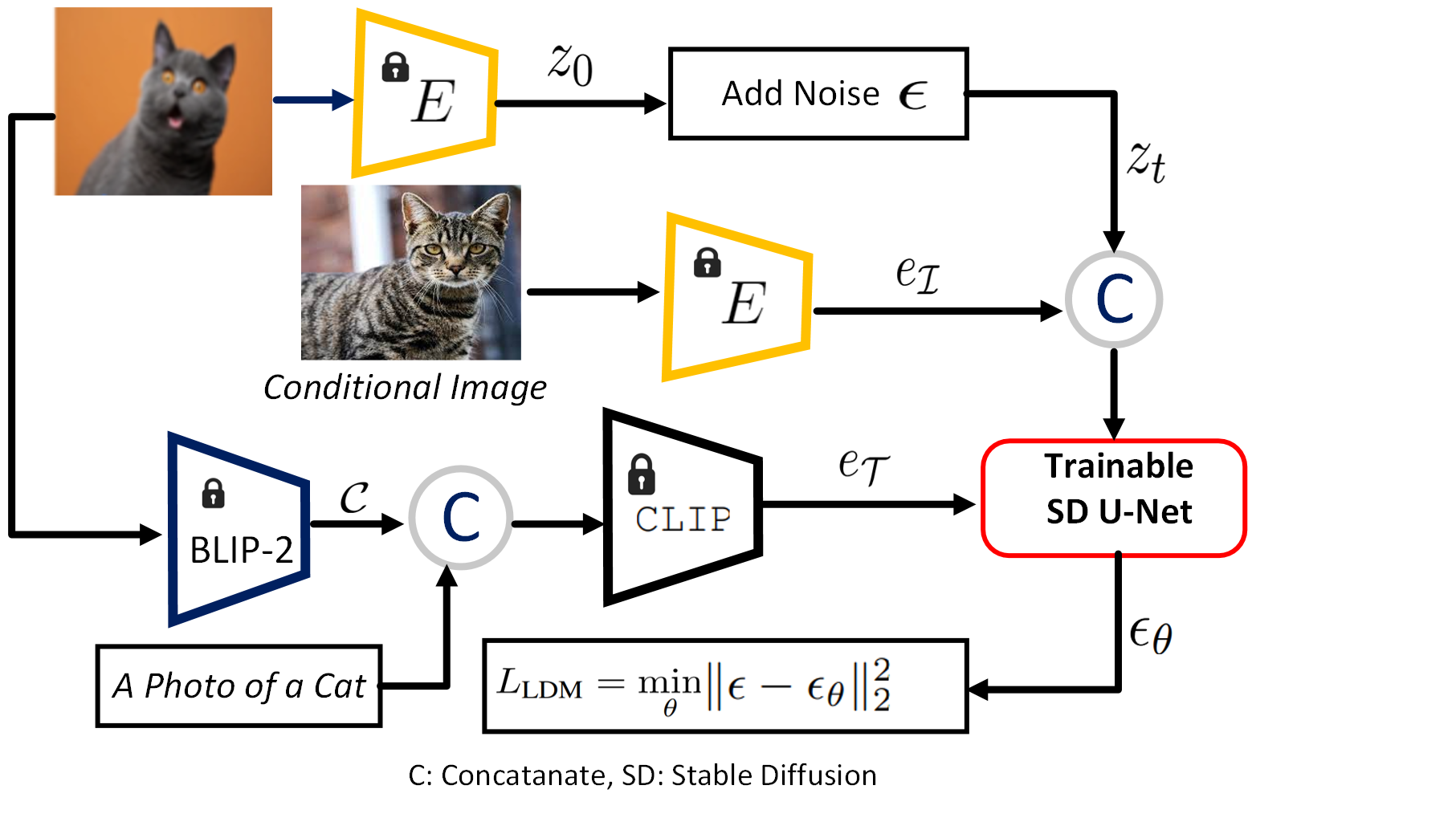}
  \vspace{-18pt}
  \caption{The training procedure illustrated with an image of a Cat: We fine-tune the pre-trained Stable Diffusion (SD U-Net) model. Input `Cat' image is used to generate noisy latent $z_t$. The `Conditional Image' is used to generate image conditioning $e_{\cI}$ which is concatenated with $z_t$. Input image caption generated by BLIP-2, is concatenated with a user-provided prompt like for e.g., ``a photo of a Cat''. The captions of all `Cat' images are stored in $\cC$ and are used to generate text conditioning $e_{\cT}$ with the MVC algorithm. The SD U-Net is trained using the objective in Equation \ref{Eq:LDM}. 
  }
  \label{Fig:train}
  \vspace{-15pt}
\end{figure}

\noindent \textbf{Training}: Figure \ref{Fig:train} illustrates the fine-tuning process of the stable diffusion model with the training dataset. During training two images $x$ and $x'$ are selected at random from the same category $y$, for e.g., `Cat'. Their embeddings are estimated using a pre-trained encoder with $z_0 \gets E(x)$ and $e_{\cI} \gets E(x')$. 
$z_0$ undergoes diffusion to yield a diffused latent $z_t$ for $t\in\{1,2,\ldots,T\}$, according to Equation \ref{Eq:diff1}. 
The noise vector $\epsilon \sim \cN(0,I)$ in Equation \ref{Eq:diff1} is the size of the image latent $z_0$.  
The other embedding $e_{\cI}$ will play the role of image conditioning and will be concatenated with $z_t$.  
The pool of captions $\cC$ for the category $y$ are used to generate new CLIP embeddings $\hat{\cE}$ using Algorithm \ref{Alg:1}. 
These embeddings will be used as the text conditioning. 
$e_{\cT} \in \hat{\cE}$, $e_{\cI}$ and $z_t$ are used as input to the U-Net of the LDM $\epsilon_{\theta}(z_t, t, e_{\cT},e_{\cI})$ to predict the noise $\epsilon_\theta$ as per Equation \ref{Eq:LDM} and train the parameters $\theta$ of the LDM using gradient descent. 
This procedure is repeated for images from every category and for all diffusion levels $t\in\{1,2,\ldots,T\}$.  \\

\begin{figure}[!t]
  \centering
  \includegraphics[height=6.3cm, width=8.5cm]{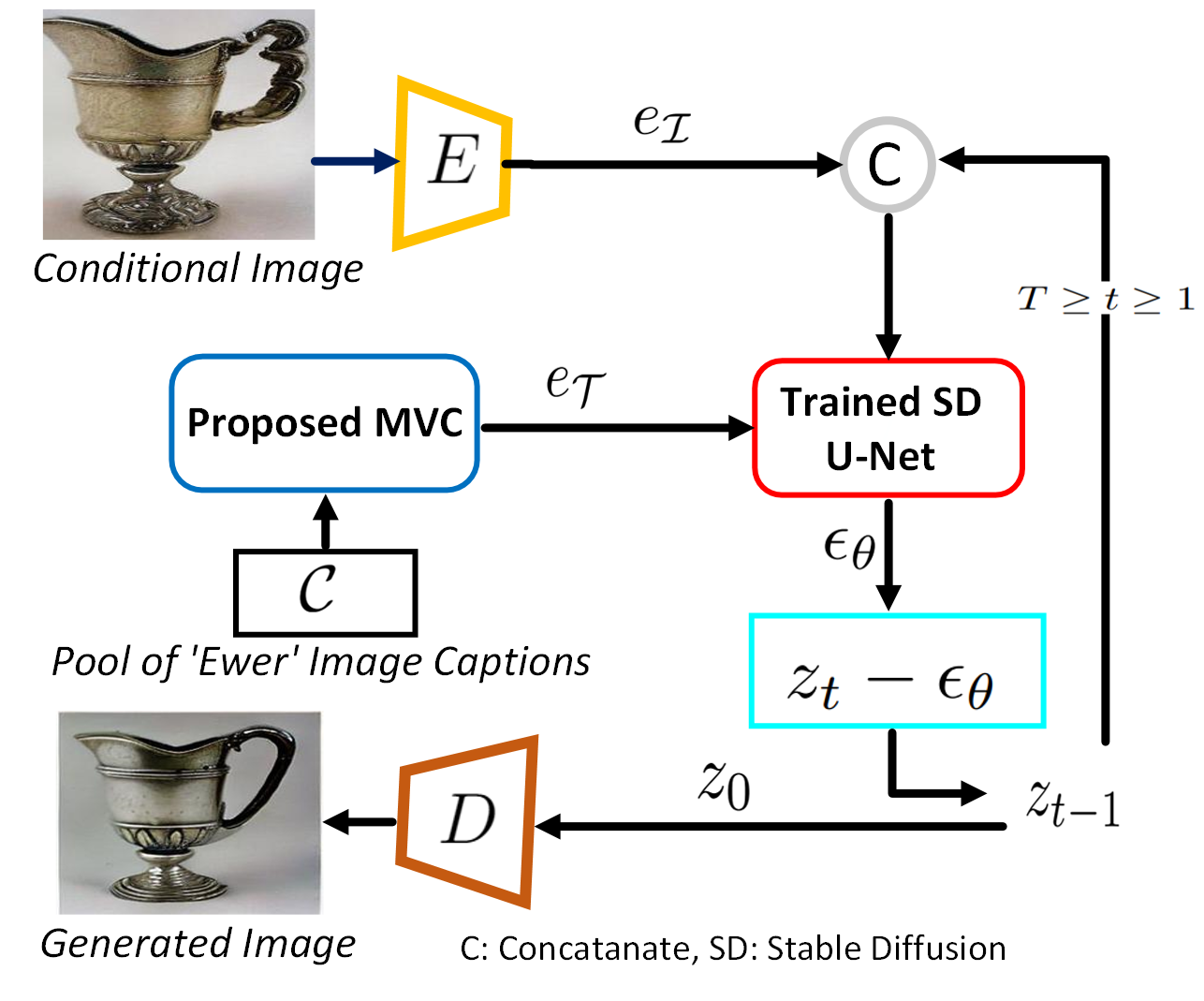}
  \vspace{-18pt}
  \caption{An overview of image generation: We begin with a complete noisy latent $z_T \sim \cN(0,I)$. To generate an `Ewer'-like image we use the Conditional Image to generate an image embedding $e_{\cI}$ and concatenate it with $z_T$. We apply the MVC algorithm on the pool of image captions of `Ewer' to obtain the text conditioning $e_{\cT}$. We apply the denoising procedure in Equation \ref{Eq:denoise} to estimate $z_{T-1}$ from $z_T$. We iterate this procedure for $T\geq t\geq 1$ to arrive at $z_0$ - the denoised latent. We apply the decoder to obtain the new image $\hat{x} \gets D(z_0)$. 
  }
  \label{Fig:Generation}
  \vspace{-15pt}
\end{figure}

\noindent \textbf{Generation}: 
The training yields the denoiser model $\epsilon_{\theta}(.)$ which can predict the noise added to the latent image representation. 
During Generation, a noise vector $z_T \sim \cN(0,I)$ is first sampled. 
To generate a new image for category $y$, for e.g., `Ewer', an image $x'$ from category $y$ is selected at random and its embedding is obtained using the Encoder $e_{\cI} \gets E(x')$. 
$e_{\cI}$ will play the role of image conditioning and will be concatenated with $z_t$. 
A new embedding $e_{\cT} \in \hat{\cE}$ is sampled to play the role of text conditioning. 
$e_{\cT}$, $e_{\cI}$ and $z_T$ are used as input to the trained U-Net of the LDM $\epsilon_{\theta}(z_T, t, e_{\cT},e_{\cI})$ to predict the noise as per Equation \ref{Eq:LDM}. 
The predicted noise $\epsilon_\theta$ is used to estimate the denoised latent $z_{T-1}$ using Equation \ref{Eq:denoise}. 
This procedure is iterated for $T\geq t\geq 1$ using the same embeddings $e_{\cT}$ and $e_{\cI}$ to obtain the completely denoised latent $z_0$. 
The pre-trained Decoder is used to generate the new image from the denoised latent with $\hat{x} \gets D(z_0)$.  Figure \ref{Fig:Generation} outlines the procedure for synthesizing new samples in a dataset.\\

Besides generating samples from our proposed methods, we use the pre-trained text-to-image diffusion model to generate new samples using the caption of images from the training dataset. For special domains like medical datasets, relying solely on pre-trained models may not be effective. However, for more general-purpose datasets such as  ImageNet \cite{deng2009imagenet}, pre-trained models can provide additional diversity and variability in generated samples. For CIFAR-10 and CIFAR-100, we generate additional images using a pre-trained Stable Diffusion (SD) model. 
\

\section{Experiments and Results}
\textbf{Stable Diffusion.} Stable Diffusion (SD) \cite{rombach2022high} is a cutting-edge text-to-image model that generates photorealistic images from text prompts. It is a conditional diffusion model, progressively refining noisy samples into realistic images that match the input text's visual context. We adopt the pre-trained SD model for text-to-image generation. We fine-tune its U-Net component to accept images from the training dataset and a prompt as conditioning inputs.

Our image augmentation process combines both our fine-tuned and a pre-trained SD model, effectively enhancing data for general classification tasks. However, for specialized domains such as medical image datasets (e.g., MRI, X-ray, CT scans), we rely exclusively on our trained SD model. This choice is due to limitations in pre-trained models' ability to accurately generate medical images, which often lack sufficient representation of medical data in their training sets. Our trained SD model excels in producing new medical images with diverse variations. To generate these images, we employ the proposed MVC procedure to generate novel text embeddings using image captions as prompts, which integrates concepts and contexts from multiple images via text embedding. Algorithm \ref{Alg:1} outlines the MVC procedure.

\begin{table*}[t]
\vspace{-12pt}
  \caption{Test accuracy (\%) on Tiny ImageNet \cite{le2015tiny} CIFAR-10, CIFAR-100 and their reduced sets. Comparisons across default data augmentation (baseline), and Fast AutoAugment (Fast AA) \cite{lim2019fast}, AutoAugment (AA) \cite{cubuk2019autoaugment} and RandAugment (RA) \cite{cubuk2020randaugment} and proposed approach (Ours). We reuse baseline results from the AA and RA paper, and results marked with * are reproduced using the same settings. The Results reported by us are averaged over 3 independent runs. The green color indicates the percent improvement from the corresponding baseline. 
  }
  \label{Tab:CIFAR}
  \centering
  \addtolength{\tabcolsep}{-0.2em}
  \begin{tabular}{@{}l|lll@{}l@{}llll@{}}

    Model & Baseline & FAA & AA & RA &  Ours(Syn) & Ours(FAA+Syn) & Ours(AA+Syn) & Ours(RA+Syn)\\
     \hline
\textbf{CIFAR-10} &                          & & & & &    \\
     
    \begin{tabular}[c]{@{}l@{}} Wide-ResNet-28-2\end{tabular} 
    & \multicolumn{1}{c}{94.9} & \multicolumn{1}{c}{$\text{95.9}^*$} & \multicolumn{1}{c}{95.9} & \multicolumn{1}{c}{95.8} &
     \multicolumn{1}{c}{95.6(\textcolor{green}{+0.7})} &
    \multicolumn{1}{c}{96.2(\textcolor{green}{+0.3})} &
    \multicolumn{1}{c}{96.3(\textcolor{green}{+0.4})} 
    & \multicolumn{1}{c}{96.6(\textcolor{green}{+0.8})} \\

   \begin{tabular}[c]{@{}l@{}} Wide-ResNet-28-10\end{tabular}  
   & \multicolumn{1}{c}{96.1} & \multicolumn{1}{c}{97.3} & \multicolumn{1}{c}{97.4} & \multicolumn{1}{c}{97.3} &
   \multicolumn{1}{c}{96.5(\textcolor{green}{+0.4})} &
   \multicolumn{1}{c}{97.6(\textcolor{green}{+0.3})} & \multicolumn{1}{c}{97.6(\textcolor{green}{+0.2})} & \multicolumn{1}{c}{97.7(\textcolor{green}{+0.4})} \\
   \hline
 \textbf{CIFAR-10 Reduced} &                          & & & & &    \\  
 \begin{tabular}[c]{@{}l@{}} Wide-ResNet-28-2\end{tabular} & \multicolumn{1}{c}{$\text{79.5}^*$} & \multicolumn{1}{c}{$\text{83.8}^*$} & \multicolumn{1}{c}{$\text{83.7}^*$} & \multicolumn{1}{c}{$\text{83.9}^*$} & 
 \multicolumn{1}{c}{82.9(\textcolor{green}{+3.4})} &
 \multicolumn{1}{c}{86.2(\textcolor{green}{+2.4})} & \multicolumn{1}{c}{86.6(\textcolor{green}{+2.9})} & \multicolumn{1}{c}{86.4(\textcolor{green}{+2.5})}
 
 \\
 \begin{tabular}[c]{@{}l@{}} Wide-ResNet-28-10\end{tabular} & \multicolumn{1}{c}{81.2} & \multicolumn{1}{c}{$\text{85.9}^*$} & \multicolumn{1}{c}{86.1} & \multicolumn{1}{c}{$\text{85.6}^*$} & 
 \multicolumn{1}{c}{85.1(\textcolor{green}{+3.9})} &
 \multicolumn{1}{c}{87.4(\textcolor{green}{+1.5})} & \multicolumn{1}{c}{87.3(\textcolor{green}{+1.2})} & \multicolumn{1}{c}{87.5(\textcolor{green}{+1.9})} \\

 \hline
\textbf{CIFAR-100} &                          & & & & &    \\

   \begin{tabular}[c]{@{}l@{}} Wide-ResNet-28-2\end{tabular} & \multicolumn{1}{c}{75.4} & \multicolumn{1}{c}{$\text{78.7}^*$} & \multicolumn{1}{c}{78.5} & \multicolumn{1}{c}{78.3} & 
   \multicolumn{1}{c}{76.9(\textcolor{green}{+1.5})} &
   \multicolumn{1}{c}{80.7(\textcolor{green}{+2.0})} & \multicolumn{1}{c}{80.4(\textcolor{green}{+1.9})} & \multicolumn{1}{c}{81.5(\textcolor{green}{+3.2})} \\

   \begin{tabular}[c]{@{}l@{}} Wide-ResNet-28-10\end{tabular} & \multicolumn{1}{c}{81.2} & \multicolumn{1}{c}{82.7} &
   \multicolumn{1}{c}{82.4} &
   \multicolumn{1}{c}{82.9} & \multicolumn{1}{c}{83.3(\textcolor{green}{+2.1})} & \multicolumn{1}{c}{84.5(\textcolor{green}{+1.8})} & \multicolumn{1}{c}{84.4(\textcolor{green}{+1.5})} & \multicolumn{1}{c}{84.9(\textcolor{green}{+1.6})} \\
   
   \hline
\textbf{CIFAR-100 Reduced} &                          & & & & &    \\

   \begin{tabular}[c]{@{}l@{}} Wide-ResNet-28-2\end{tabular} & \multicolumn{1}{c}{$\text{39.7}^*$} & \multicolumn{1}{c}{$\text{45.4}^*$} & \multicolumn{1}{c}{$\text{44.7}^*$} & \multicolumn{1}{c}{$\text{44.8}^*$} & 
   \multicolumn{1}{c}{46.1(\textcolor{green}{+6.4})} &
   \multicolumn{1}{c}{47.7(\textcolor{green}{+2.3})} & \multicolumn{1}{c}{47.5(\textcolor{green}{+2.8})} & \multicolumn{1}{c}{47.8(\textcolor{green}{+3.0})} \\
   
   \begin{tabular}[c]{@{}l@{}} Wide-ResNet-28-10\end{tabular} & \multicolumn{1}{c}{$\text{41.5}^*$} & \multicolumn{1}{c}{$\text{46.3}^*$} & \multicolumn{1}{c}{$\text{45.9}^*$} & \multicolumn{1}{c}{$\text{45.7}^*$} & 
   \multicolumn{1}{c}{47.8(\textcolor{green}{+6.3})} &
   \multicolumn{1}{c}{48.7(\textcolor{green}{+2.4})} & \multicolumn{1}{c}{48.1(\textcolor{green}{+2.2})}& \multicolumn{1}{c}{48.5(\textcolor{green}{+2.8})} \\
   
  \hline

\textbf{Tiny ImageNet} &                          & & & & &    \\

   \begin{tabular}[c]{@{}l@{}} ResNet50\end{tabular} & \multicolumn{1}{c}{$\text{27.3}^*$} & \multicolumn{1}{c}{$\text{30.5}^*$} & \multicolumn{1}{c}{$\text{30.4}^*$} & \multicolumn{1}{c}{-} & 
   \multicolumn{1}{c}{33.2(\textcolor{green}{+5.9})} &
   \multicolumn{1}{c}{35.4(\textcolor{green}{+2.3})} & \multicolumn{1}{c}{35.5(\textcolor{green}{+2.8})} & \multicolumn{1}{c}{-} \\
   
   \begin{tabular}[c]{@{}l@{}} EfficientNet-b3\end{tabular} & \multicolumn{1}{c}{$\text{46.3}^*$} & \multicolumn{1}{c}{$\text{-}$} & \multicolumn{1}{c}{$\text{-}$} & \multicolumn{1}{c}{$\text{47.9}^*$} & 
   \multicolumn{1}{c}{48.6(\textcolor{green}{+2.3})} &
   \multicolumn{1}{c}{-} & \multicolumn{1}{c}{-}& \multicolumn{1}{c}{49.4(\textcolor{green}{+2.8})} \\

  \hline

  \begin{tabular}[c]{@{}l@{}} Average \end{tabular} & \multicolumn{1}{c}{-} & \multicolumn{1}{c}{-} & \multicolumn{1}{c}{-} & \multicolumn{1}{c}{-} & \multicolumn{1}{c}{(\textcolor{green}{+3.29})} & \multicolumn{1}{c}{(\textcolor{green}{ 
    +1.70})} & \multicolumn{1}{c}{(\textcolor{green}{+1.77})} & \multicolumn{1}{c}{(\textcolor{green}{+2.11})} \\
  \end{tabular}
 
\end{table*}

\begin{table*}[tb]
  \caption{Test accuracy (\%) on Caltech101 datasets \cite{fei2006one}. Comparisons across default data augmentation (baseline), AutoAugment (AA) \cite{cubuk2019autoaugment}, RandAugment (RA) \cite{cubuk2020randaugment} and proposed approach (Ours). All of the Results are reported by us averaging over 3 independent runs. The green color indicates the percent improvement from the corresponding baseline.
  }
  \vspace{-10pt}
\label{Tab:Caltech101}
  \centering
  \begin{tabular}{@{}l|llllll@{}}

    Model & Baseline & AA &  RA & Ours(AA+Syn) & Ours(RA+Syn) \\
     \hline

     \begin{tabular}[c]{@{}l@{}} RestNet50\end{tabular} 
     & \multicolumn{1}{c}{95}
     & \multicolumn{1}{c}{96.1}
      & \multicolumn{1}{c}{96.4} 
    & \multicolumn{1}{c}{97.2(\textcolor{green}{+1.1})}
   
    & \multicolumn{1}{c}{97.5(\textcolor{green}{+1.1})} & 

\\
    \begin{tabular}[c]{@{}l@{}} EfficientNet-b0\end{tabular} & \multicolumn{1}{c}{95.2} & 
    \multicolumn{1}{c}{95.1} & 
      \multicolumn{1}{c}{95.5} &
    \multicolumn{1}{c}{95.8(\textcolor{green}{+0.7})} &
   
    \multicolumn{1}{c}{96.3(\textcolor{green}{+0.8})} &  \\

   \begin{tabular}[c]{@{}l@{}} VIT-16\end{tabular}
   & \multicolumn{1}{c}{96.2}
   & \multicolumn{1}{c}{96.7}
    & \multicolumn{1}{c}{96.9} 
   & \multicolumn{1}{c}{97.5(\textcolor{green}{+0.8})}
   & \multicolumn{1}{c}{97.9(\textcolor{green}{+1.0})} &  \\
   
  \hline
   \begin{tabular}[c]{@{}l@{}} Average \end{tabular}
   & \multicolumn{1}{c}{-}
   & \multicolumn{1}{c}{-}
    & \multicolumn{1}{c}{-(\textcolor{green}{+0.3})} 
   & \multicolumn{1}{c}{(\textcolor{green}{+0.87})}
   & \multicolumn{1}{c}{(\textcolor{green}{+0.97})}
  \end{tabular}
  \vspace{-10pt}
\end{table*}

\begin{table*}[tb]
\vspace{-10pt}
  \caption{Test accuracy (\%) on Brain Tumor Dataset \cite{1jny-g144-23}. Comparisons across, AutoAugment (AA)  \cite{cubuk2019autoaugment}, RandAugment(RA) \cite{cubuk2020randaugment}  and and proposed approach (Ours). All of the Results are reported by us averaging over 3 independent runs. The green color indicates the percent improvement from the corresponding baseline.
  }
  \label{Tab:Medical}
  \centering
  \begin{tabular}{@{}l|llllll@{}}

    Model & Baseline & AA  & RA &  Ours(AA+Syn)  & Ours(RA+Syn) \\
     \hline

     \begin{tabular}[c]{@{}l@{}} RestNet50\end{tabular} & \multicolumn{1}{c}{93.2
} & \multicolumn{1}{c}{94.4} 
& \multicolumn{1}{c}{95.6}
& \multicolumn{1}{c}{96.1(\textcolor{green}{+1.7})}

& \multicolumn{1}{c}{96.8(\textcolor{green}{+1.2})}&  \\
   
    \begin{tabular}[c]{@{}l@{}} EfficientNet-B0\end{tabular} & \multicolumn{1}{c}{91.9} & \multicolumn{1}{c}{93.3}   & \multicolumn{1}{c}{94.5} & \multicolumn{1}{c}{95.2 (\textcolor{green}{+1.9})}  & \multicolumn{1}{c}{95.2(\textcolor{green}{+0.7})} &  \\

   \begin{tabular}[c]{@{}l@{}} Wide-RestNet-50-2\end{tabular} & \multicolumn{1}{c}{92.3} & \multicolumn{1}{c}{94.7} & \multicolumn{1}{c}{95.6} & \multicolumn{1}{c}{95.8(\textcolor{green}{+1.1})}&  \multicolumn{1}{c}{96.7(\textcolor{green}{+1.1})} &  \\
   
  \hline

   \begin{tabular}[c]{@{}l@{}} Average \end{tabular}
   & \multicolumn{1}{c}{-}
   & \multicolumn{1}{c}{-}
    & \multicolumn{1}{c}{-} 
   & \multicolumn{1}{c}{(\textcolor{green}{+1.43})}
   & \multicolumn{1}{c}{(\textcolor{green}{+1.13})} &  
  \end{tabular}
  \vspace{-10pt}
\end{table*}

\textbf{Datasets.}
To explore the space of data augmentations, we experiment with core image classification datasets such as CIFAR-10 and CIFAR-100 \cite{krizhevsky2009learning}, which are commonly used benchmarks in the field. These datasets enable direct comparisons with prior research. We also consider Tiny ImageNet \cite{le2015tiny}, one of the harder datasets in the classification task. In addition, we include medical image datasets \cite{1jny-g144-23} to broaden our experimentation scope, highlighting the benefits of augmented images in training compared to traditional augmentation methods. Furthermore, we explore the Caltech-101 dataset \cite{fei2006one}, renowned for its fine-grained classification challenges. We reorganize the training dataset by randomly selecting pairs of images with the same class label and pairing them to fine-tune the SD model. Details of data preparation are discussed in the Supplementary.

\textbf{Training.}   For fine-tuning the SD model, we adopt an end-to-end training approach inspired by InstructPix2Pix \cite{brooks2023instructpix2pix} using our curated source dataset. Similar to current augmentation techniques\cite{cubuk2019autoaugment,lim2019fast, cubuk2020randaugment}, the training of a classification model for CIFAR-10 and CIFAR-100 datasets starts from scratch with similar hyperparameter settings. Tiny ImageNet: we choose a learning rate of $10^{-3}$ and 50 epochs for ResNet50. For EfficientNet-B3, we use a learning rate of $10^{-2}$ and 20 epochs. We select a batch size of 128 for both models.  Caltech101: we choose a learning rate of $10^{-4}$ with an ADAM optimizer. For Brain Tumor Dataset, different hyperparameter settings are selected for various classification models, with details provided in the Supplementary.  

\subsection{Results}
\noindent \textbf{CIFAR-10 and CIFAR-100 Results.}
In Table \ref{Tab:CIFAR}, we present the test set accuracy for different neural network architectures. We implemented the Wide-ResNet-28-2 and Wide-ResNet-28-10 \cite{Zagoruyko2016Wide} models in PyTorch and used the same model and hyperparameter settings to evaluate the test set accuracy. As shown in the Table \ref{Tab:CIFAR}, we achieved an average accuracy improvement of 3.29\%, 1.70\%, 1.77\%, and 2.11\% compared to Baseline, FastAutoAugment, AutoAugment, and RandAugment respectively. Similar to AutoAugment, we trained on $4,000$ labeled examples randomly sampled from the CIFAR-10 dataset, which we referred to as CIFAR-10 Reduced and CIFAR-100 Reduced. 

\noindent \textbf{Fine-grained Classification Result.}
The effectiveness of diffusion model augmentation in the fine-grained domain was investigated using the Caltech101 dataset \cite{fei2006one}. This dataset presents a challenge due to its imbalanced nature and relatively small training sets, despite containing numerous classes. In Table \ref{Tab:Caltech101}, the classification results show that our methods outperformed the baseline, AutoAugment and RandAugment by a significant margin.

\noindent \textbf{Medical Image Dataset Results.} 
Medical images differ significantly from images in other fields. Moreover, obtaining real MRI or CT scan datasets can be challenging due to the high cost and limited availability of patients. Therefore, data augmentation is a common necessity in the medical field. We assessed our proposed MVC method for augmentation in the medical domain and the classification accuracy results are presented in Table \ref{Tab:Medical}, showing a $1.43\%$ and $1.13\%$  improvement compared to state-of-the-art methods. We note that as the size of the training set increases, the impact of data augmentation is expected to diminish. We observed that for CIFAR-10 and the reduced set CIFAR-100, we achieved a greater improvement in percentages compared to the full set.

\subsection{Ablation Study}
\subsubsection{Is Diversity Enough?}

\begin{figure}
\centerline{\includegraphics[width=8.5cm ,  height= 5cm]{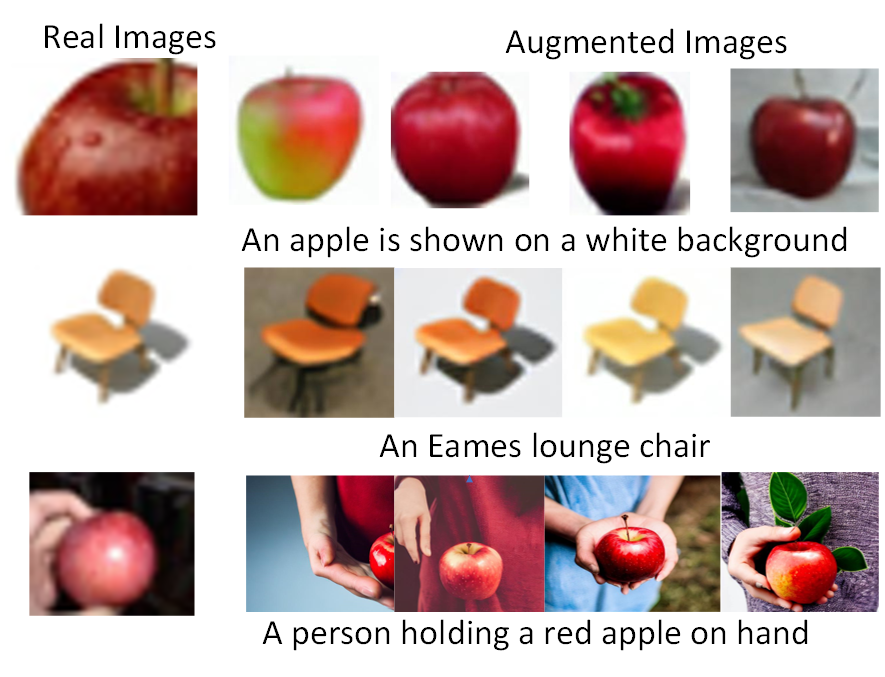}}
\vspace{-10pt}
\caption{Column 1 depicts real images from CIFAR-100. In the remaining columns, the first two rows are images generated using the proposed MVC method, which constrains generated images to be in-domain. The third row represents images generated using a pre-trained SD model, which provides more diversity with no control over in-domain generation.}
\label{Fig:examples2}
\vspace{-10pt}
\end{figure}

Our study investigated the impact of image data augmentation on classification accuracy. We observed that enhancing diversity alone doesn't consistently improve classification results unless out-of-domain sample generation is carefully managed. Specifically, when utilizing pre-trained diffusion models, such as stable diffusion, we found that the samples generated often deviated significantly from the dataset's domain. This effect was particularly pronounced in fine-grained datasets like Caltech101 and specialized domains such as MRI, X-ray, and CT-SCAN. Figure \ref{Fig:Fig1} illustrates the limitations of pre-trained SD models in the medical domain. However, there are cases where pre-trained SD models can produce diverse samples that align well with the dataset. For instance, in Figure \ref{Fig:examples2}, the last row demonstrates diversified augmented samples derived from CIFAR-100.

\begin{table}[tb]
  \caption{Impact of Pre-trained (PT) vs fined-tuned (FT) SD model on test accuracy (\%) on Caltech101, CIFAR100, and Brain Tumor Dataset. Comparisons across AutoAugment (AA)\cite{cubuk2019autoaugment} + Synthetic Data from  Pre-trained \cite{rombach2022high} and Fined-tuned SD Model \cite{brooks2023instructpix2pix}.}
  
  \label{tab:impact}
  \centering
  \addtolength{\tabcolsep}{-0.3em}
  \begin{tabular}{@{}l|llll@{}}

    Model & AA  &  AA+Syn(PT)  &  AA+Syn(FT)  \\
     \hline
\textbf{Caltech101} &                          & & &    \\
     \begin{tabular}[c]{@{}l@{}} RestNet50\end{tabular} & \multicolumn{1}{c}{96.1
} 
& \multicolumn{1}{c}{96.8(\textcolor{green}{+0.7})}
&
 \multicolumn{1}{c}{97.2(\textcolor{green}{+1.1})} &  \\
    \begin{tabular}[c]{@{}l@{}} EfficientNet-B0\end{tabular} & \multicolumn{1}{c}{95.1} &  \multicolumn{1}{c}{95.6(\textcolor{green}{+0.5})} &  \multicolumn{1}{c}{95.8(\textcolor{green}{+0.7})} & \\
    
   \begin{tabular}[c]{@{}l@{}} VIT-16\end{tabular}  & \multicolumn{1}{c}{96.7} & \multicolumn{1}{c}{97.3(\textcolor{green}{+0.6})}  & \multicolumn{1}{c}{97.5(\textcolor{green}{+0.8})}&   \\
   
  \hline
\textbf{CIFAR100 Reduced} &                          & & &    \\
  \begin{tabular}[c]{@{}l@{}} Wide-ResNet-28-2\end{tabular}  & \multicolumn{1}{c}{44.7} & \multicolumn{1}{c}{46.1(\textcolor{green}{+01.4})}  & \multicolumn{1}{c}{47.3(\textcolor{green}{+2.6})}&   \\
\hline
\textbf{Brain Tumor Dataset} &                          & & &    \\
  \begin{tabular}[c]{@{}l@{}} ResNet50\end{tabular}  & \multicolumn{1}{c}{94.4} & \multicolumn{1}{c}{-}  & \multicolumn{1}{c}{96.1(\textcolor{green}{+1.7})}&   \\
  \hline
   \begin{tabular}[c]{@{}l@{}} Average\end{tabular}  & \multicolumn{1}{c}{-} & \multicolumn{1}{c}{(\textcolor{green}{+0.8})}  & \multicolumn{1}{c}{(\textcolor{green}{+1.38})}&  
  \end{tabular}
  \vspace{-5pt}
\end{table}

\begin{figure}
\centerline{\includegraphics[width=8.5cm ,  height= 5cm]{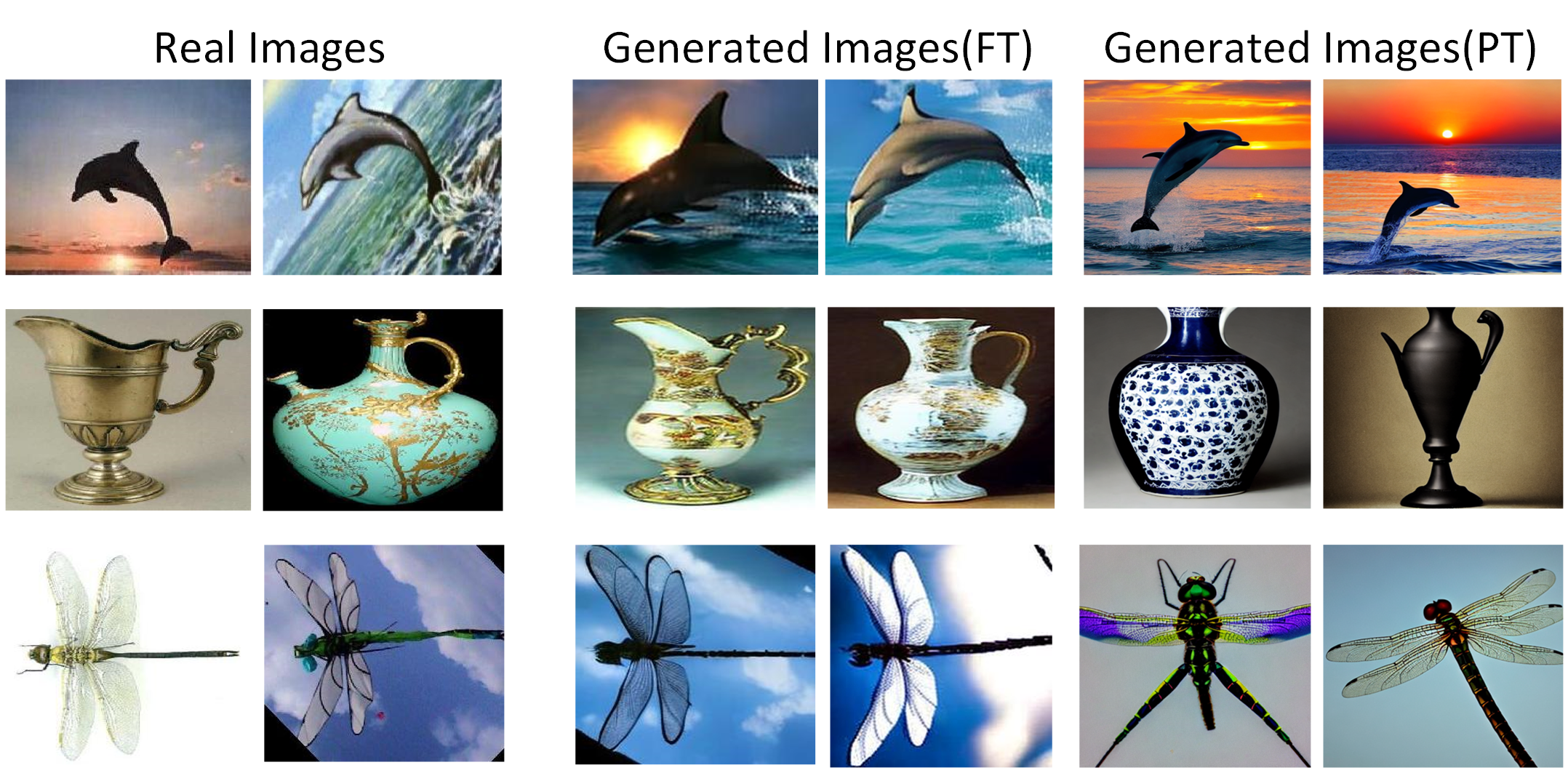}}
\vspace{-10pt}
\caption{ The impact of concept mixing on augmented images that were created using the MVC method. The first two images on the left side represent the original images from the Caltech101 dataset, while the subsequent images are the generated ones from fine tuned DM and pretrained DM.}
\label{Fig:examples}
\vspace{-20pt}
\end{figure}

\noindent \textbf{Need for Fine-Tuning and Guided Augmentation.} To address pre-trained model challenges, our research emphasizes fine-tuning diffusion models for fine-grain and medical imaging datasets. Fine-tuning allows models to adjust generation capabilities to better match dataset characteristics, improving sample quality and relevance. We stress the importance of a tailored augmentation prompting algorithm to guide sample generation, aligning outputs with domain-specific features. Table \ref{tab:impact} illustrates the effect of using a pre-trained and fine-tuned SD model on test accuracy, and it also presents the average accuracy improvement for the fine-tuned setting. Additionally, Figure \ref{Fig:examples} illustrates how concept mixing influences the diversity of augmented images created using the proposed MVC method. It combines color and shape concepts from multiple real images into augmented images.  However, generated images from a pretrained model look more diverse, but they exhibit some moderate deviation from in domain distribution which may affect model learning.

\subsubsection{Effective Training Strategy} 
\textbf{Learning from Image-Caption Pairs.} We initially trained the diffusion model with paired image-caption data. This method focused on teaching the model to generate images based on specific textual descriptions of images. While effective in reinforcing direct associations between images and captions, its ability to generate diverse outputs beyond paired examples was limited.

\noindent \textbf{Generating Diverse In-domain Samples.} In the second approach, we created a dataset pairing images and captions randomly selected from the same class label. This method aimed to broaden the model's capacity to generate diverse images within the context of specific class labels. This approach yielded superior results compared to the first method. It enhanced the model's ability to produce varied but contextually relevant images within defined domains.

\noindent \textbf{Impact of Synthetic Data Quantity.} In our exploration of integrating augmented synthetic datasets to train classification models, we tested three different approaches. Initially, we found that simply adding the synthetic dataset to the real training dataset did not significantly boost classification accuracy. This approach proved ineffective, especially when the synthetic dataset outnumbered the real dataset by a large margin, leading to a noticeable bias towards the synthetic data. This bias occurred because the model had more exposure to synthetic examples, potentially overshadowing the patterns present in the real data. Later we tested two other strategies. two-phase training and  Random Selection with Probability (RSP). 

\noindent \textbf{Two-phase Training:} In the first phase, we trained the classification models using a combined dataset. This dataset included both real and synthetic data, with the synthetic data being three to four times larger than the real data. During this phase, the model focused on learning general patterns and features present in both types of data. Exposure to a large volume of synthetic data helped the model understand the wider range of possible inputs. After the initial training phase, we fine-tuned the model using only the real dataset, with reduced training steps and a smaller learning rate. This phase aimed to refine the model's understanding specifically on real-world examples. By limiting the number of training iterations during fine-tuning, we prevented the model from overfitting to the real dataset.  

\noindent \textbf{Random Selection with Probability (RSP):} This method involved randomly selecting synthetic images with a specified probability (e.g., 80\%) from a pool of synthetic datasets and integrating them with batches from the real dataset during training. This approach aimed to dynamically balance the integration of synthetic and real data to enhance training effectiveness. Table \ref{Tab:abs5} compares the different training approaches where we see two-phase training outperformed others. Training strategies are given in the Supplementary.

\begin{table}[tb]
  \caption{Impact of different training strategies on test accuracy (\%) on Caltech101 Dataset.Comparisons across AutoAugment (AA) \cite{cubuk2019autoaugment}, combined (real+synthetic), Random Selection with Probalibity(RSP) and two-phase training strategies.
  }
  \vspace{-10pt}
  \label{Tab:abs5}
  \centering
  \begin{tabular}{@{}l|lllll@{}}

    Model & AA  &  combined  & RSP & Two-Phase \\
     \hline

     \begin{tabular}[c]{@{}l@{}} RestNet50\end{tabular} & \multicolumn{1}{c}{96.1
} 
& \multicolumn{1}{c}{95.4(\textcolor{red}{-0.7})}
& \multicolumn{1}{c}{96.3} &
 \multicolumn{1}{c}{97.2(\textcolor{green}{+1.1})} &  \\
    \begin{tabular}[c]{@{}l@{}} EfficientNet-B0\end{tabular} & \multicolumn{1}{c}{95.1} &  \multicolumn{1}{c}{94.9(\textcolor{red}{-0.2})} & \multicolumn{1}{c}{95.3} & \multicolumn{1}{c}{95.8(\textcolor{green}{+0.7})} & \\
    
   \begin{tabular}[c]{@{}l@{}} VIT-16\end{tabular}  & \multicolumn{1}{c}{96.7} & \multicolumn{1}{c}{97.2(\textcolor{green}{+0.5})}  & \multicolumn{1}{c}{97.2}& \multicolumn{1}{c}{97.5(\textcolor{green}{+0.8})}&   \\
  \hline
  
  \end{tabular}
  \vspace{-10pt}
\end{table}

\section{Conclusion}
Our experiments demonstrate that synthetic data significantly improves classifier learning. 
However, we identified limitations with pre-trained diffusion models. 
Despite their general capability to produce high-quality samples, pre-trained diffusion models struggle when it comes to aligning with specific dataset domains, such as fine-grained classification and medical images, likely due to inherent training biases. This lack of alignment can lead to sub-optimal performance and decreased effectiveness in these specialized domains. To address these limitations, we emphasize the critical role of fine-tuning and domain-specific augmentation strategies in mitigating out-of-domain sample generation issues.  Our proposed fine-tuning strategies and MVC method are crucial to enhance the quality and applicability of generated samples in specialized domains, ultimately improving the effectiveness of machine learning models in real-world applications.

\section{Acknowledgment}

This research was sponsored by the Army Research Laboratory and was accomplished under
Cooperative Agreement Number W911NF-23-2-0224. The views and conclusions contained in
this document are those of the authors and should not be interpreted as representing the
official policies, either expressed or implied, of the Army Research Laboratory or the U.S.
Government. The U.S. Government is authorized to reproduce and distribute reprints for
Government purposes notwithstanding any copyright notation herein.

\newpage
{\small
\bibliographystyle{ieee_fullname}
\bibliography{main}
}

\end{document}